\documentclass[runningheads]{llncs}

\usepackage{eccv}

\usepackage{eccvabbrv}

\usepackage{graphicx}
\usepackage{booktabs}

\usepackage[accsupp]{axessibility}  % Improves PDF readability for those with disabilities.

\usepackage{hyperref}
\usepackage{multirow} 
\usepackage{orcidlink}
\usepackage{marvosym}

\usepackage{array}

\begin{document}

\title{Target-Oriented Object Grasping via Multimodal Human Guidance} 


\author{
Pengwei Xie\inst{1}\orcidlink{0000-0003-1005-9252} \and
Siang Chen\inst{1, 2}\orcidlink{0000-0002-9235-6439} \and
Yixiang Dai\inst{1}\orcidlink{0009-0000-7501-5504} \and
Dingchang Hu\inst{1}\orcidlink{0000-0002-2284-4824} \and
Kaiqin Yang\inst{1}\orcidlink{0000-0003-0232-2698} \and
Guijin Wang\inst{1.2\textsuperscript{(\Letter)}}\orcidlink{0000-0002-2131-3044}
}

\authorrunning{Xie et al.}

\institute{
Department of Electronic Engineering, Tsinghua University, Beijing 100084, China \and
Shanghai AI Laboratory, Shanghai 200232, China \\
\email{wangguijin@tsinghua.edu.cn}
}

\maketitle

\begin{abstract}
In the context of human-robot interaction and collaboration scenarios, robotic grasping still encounters numerous challenges. Traditional grasp detection methods generally analyze the entire scene to predict grasps, leading to redundancy and inefficiency. In this work, we reconsider 6-DoF grasp detection from a target-referenced perspective and propose a \textit{\textbf{T}}arget-\textit{\textbf{O}}riented \textit{\textbf{G}}rasp \textit{\textbf{Net}}work (\textit{\textbf{TOGNet}}). \textit{TOGNet} specifically targets local, object-agnostic region patches to predict grasps more efficiently. It integrates seamlessly with multimodal human guidance, including language instructions, pointing gestures, and interactive clicks. Thus our system comprises two primary functional modules: a guidance module that identifies the target object in 3D space and \textit{TOGNet}, which detects region-focal 6-DoF grasps around the target, facilitating subsequent motion planning. Through 50 target-grasping simulation experiments in cluttered scenes, our system achieves a success rate improvement of about 13.7\%. In real-world experiments, we demonstrate that our method excels in various target-oriented grasping scenarios.

\keywords{Grasp Detection \and Computer Vision \and Vision Language Models \and Target-oriented Grasping}
\end{abstract}

\section{Introduction}
\label{sec:intro}
In Human-Robot-Interaction and Collaboration (HRI/C) scenarios, a robot system is designed to identify and respond to the human behaviour and collaborate with human to achieve a common goal \cite{robinson2023robotic, semeraro2023human}. A common HRI scenario is that a robot receives the inputs from humans and picks the proper object, which is supportive for patients with disabilities to assist their activities in daily life \cite{mohebbi2020human}. To accomplish a natural HRI assisting in target oriented grasping, it is essential to understand and react to the various interactions between human and robots. 

Human communication is inherently multimodal \cite{constantin2022interactive}. Language instructions \cite{holzapfel2008dialogue, weld2022survey, constantin2022interactive, constantin2023multimodal}, hand gestures or pointing gestures \cite{cao2017realtime, medeiros20213d, constantin2022interactive, constantin2023multimodal} have been widely used to guide robotic systems interacting with human and help resolve user's intention and improve the understanding of the scene. However, robotic systems are also essential for reacting to more physical commands, such as object grasping and manipulation, to help people accomplish specific tasks.

Based on advances in deep learning, many vision-based deep networks have been designed to detect grasp poses in open-world scenes. Earlier methods \cite{morrison2018closing,kumra2020antipodal} detect 4-DoF planar grasps given a single-view RGB-D image in simple scenarios with high efficiency. However, the planar representations inherently constrain the gripper to be perpendicular to the camera plane, which is limited in complex contexts. 

Accordingly, 6-DoF grasp detection has drawn extensive attention because of its broader applications. Pioneer methods employ a sampling-evaluation strategy \cite{ten2017grasp,liang2019pointnetgpd} or a direct regression paradigm \cite{ni2020pointnet++, qin2020s4g} to predict the relevant attributes, which can be time-consuming or imprecise. Recent works \cite{zhao2021regnet,wei2021gpr,wang2021graspness,chen2023hggd} encode the global information to pinpoint areas with high ``graspability" ( (i.e., the predicted success probability of grasping), and then generate grasps. Nonetheless, they are tailored to detect grasps for the entire scene without considering the suitability for specific downstream applications. When applying target-oriented scenarios, most approaches \cite{murali20206,sundermeyer2021contact,liu2022ge, liu2024okrobot} identify the target-referenced grasps by simply using filtering according to the target segmentation or detection results. These approaches introduce unnecessary computational load and risk interference from irrelevant objects, potentially leading to low-quality grasps or no grasp in the target area. Also, the quality of the results still needs to be improved, especially when facing novel scenes. 

\begin{figure}[t] 
\centering
{\includegraphics[width=12.5cm]{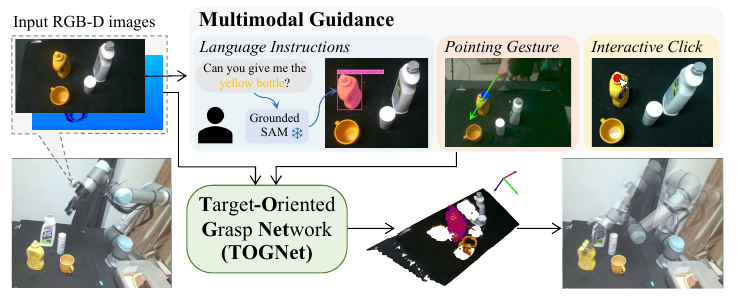}}
\caption{Our \textit{\textbf{T}}arget-\textit{\textbf{O}}riented \textit{\textbf{G}}rasp \textit{\textbf{Net}}work (\textit{\textbf{TOGNet}}) is designed to integrate seamlessly with various forms of multimodal human guidance. By cropping and analyzing the RGB-D information within the target area, \textit{TOGNet} detects high-quality 6-DoF grasps. Then the robot selects the most suitable grasp for execution. The integrated system enables applications in diverse HRI/C scenarios, providing useful assistance to people with visual, auditory, or motor impairments.}
\label{fig:teaser} 
\vspace{-0.5cm} % adjust distance above distance 
\end{figure}

In this paper, we propose a system designed for target-oriented grasping that responds to humans' interactive, multimodal guidance instructions. As illustrated in Fig. \ref{fig:teaser}, building on top of recent advanced works like Grounding DINO \cite{liu2023grounding}, MobileSAM \cite{zhang2023faster}, and MediaPipe \cite{lugaresi2019mediapipe}, we construct a pipeline to parse the human's intentions from language instructions, pointing gestures and interactive clicks, and identify the target area. Then our \textit{\textbf{T}}arget-\textit{\textbf{O}}riented \textit{\textbf{G}}rasp \textit{\textbf{Net}}work (\textit{\textbf{TOGNet}}) processes the regional information to generate grasps. We choose the best grasp pose and execute the motion planning algorithm \cite{chitta2012moveit} to control the robot picking the target object.

Specifically, our major contributions are presented as follows.
\begin{itemize}
\item We construct a pipeline to locate a target from multimodal human guidance. The system integrates multiple state-of-the-art computer vision models capable of analyzing and understanding humans' intentions and cropping the target object efficiently. The proposed system is especially helpful for assisting people with visual, auditory, or motor impairments.
\item We design a \textit{\textbf{T}}arget-\textit{\textbf{O}}riented \textit{\textbf{G}}rasp \textit{\textbf{Net}}work (\textit{\textbf{TOGNet}}), aiming to detect 6-DoF grasp poses from target-referenced regions, facilitating the motion planning process of the robot. \textit{TOGNet} is trained on a refined region-focal dataset, which is object-agnostic and thus effective when facing novel scenes.
\item We evaluate our system using datasets, simulation experiments, and real-world experiments. Specifically, we propose a new evaluation metric to adapt the target-oriented setting in our work and compare the grasp quality with recent state-of-the-art approaches. Then, we implement them in the Maniskill2 \cite{gu2023maniskill2} benchmark to assess the success rate of target-oriented grasping involving 50 cluttered scenes. Finally, we build the system in a real robot platform to verify the effectiveness of our system for understanding and reacting to multimodal human guidance. 
\end{itemize}

\section{Related Work}

Scene-level grasping involves generating grasps for the entire scene in a target-agnostic manner \cite{fang2020graspnet}. Traditional grasp detection methods typically extract semantic or geometric information from the entire scene \cite{ni2020pointnet++,qin2020s4g}. They often employ PointNet++ \cite{qi2017pointnet++} to encode scene points and directly regress grasp attributes. Recently, grasp generation has predominantly adopted a two-stage approach, focusing on \textit{where} to grasp and \textit{how} to grasp. In this context, several studies \cite{fang2020graspnet,zhao2021regnet,wang2021graspness,liu2022transgrasp,liu2023joint} extract global per-point features, assigning each point a ``graspness'' or confidence score. Then, several seed points are selected according to the scores, and grasps are detected in the nearby regions based on the features. Chen et al. \cite{chen2023hggd} enhance the feature extraction process by integrating points within graspable regions with corresponding semantic features, using a vanilla PointNet \cite{qi2017pointnet} to extract geometric and semantic features.

Conventional methods require a three-stage process in target-oriented grasping scenarios: scene-level grasping, target segmentation, and grasp filtering \cite{liu2024okrobot}. While these methods can generate many grasps for every graspable region, they often impose an excessive computational burden and do not consistently generate high-quality grasps for the intended target. In contrast, our method directly generates grasps on targeted regions and seamlessly integrates multimodal guidance techniques. 

Recent studies have utilized query images \cite{lou2021c,lou2022l,sundermeyer2021contact,liu2022ge} or language instructions \cite{xu2023a,chen2021a} to guide the grasping of specific objects in cluttered scenes. Many approaches \cite{lou2021c,lou2022l,sundermeyer2021contact,liu2022ge} integrate an additional segmentation branch to target specific objects in clutter. Xu et al. \cite{xu2023a} deploy CLIP \cite{radford2021l} for visual-language pre-processing, generating grasp poses using GraspNet \cite{fang2020graspnet}. Lu et al. \cite{lu2023vl} utilize a visual grounding algorithm for object detection, applying bounding box filters to extract object-level point clouds. These point clouds are then processed using a network \cite{lu2022hybrid} trained on scene-level data, potentially leading to suboptimal grasps due to domain mismatches between regional and scene-level geometric information. 

Moreover, recent advancements in Large Language Models (LLMs) \cite{brown2020language,touvron2023llama} and large Vision Language Models (VLMs) \cite{kirillov2023segment,liu2023grounding} have significantly enhanced robotic capabilities in complex task planning and manipulation. Xu et al. \cite{xu2023object} use GPT-3 \cite{brown2020language} to parse language instructions for object placement tasks. Yang et al. \cite{yang2023transferring} deploy Grounding DINO \cite{liu2023grounding} and SAM \cite{kirillov2023segment} to localize objects from language instructions. Liu et al. \cite{liu2024okrobot} utilize VoxelMap \cite{yenamandra2023homerobot} for object localization with natural language queries and AnyGrasp \cite{fang2023anygrasp} to detect scene-level grasp poses, subsequently filtered through object masks by LangSam \cite{Medeiros2023}. However, these methods primarily focus on interpreting language instructions and planning manipulations, often needing more efficient and high-quality grasp estimation.

Unlike them, our \textit{TOGNet} is trained on a refined region-focal dataset, efficiently generating high-quality grasps for targeted regions. This approach allows for directly processing local information rather than entire scenes, significantly reducing redundant computations. Our pipeline produces high-quality grasps that streamline the entire grasping and planning process by utilizing multimodal guidance methods such as language instructions, pointing gestures, and interactive image clicks.

\section{Method}

\subsection{Overview}

\begin{figure}[t]
\centering
\includegraphics[width=12cm]{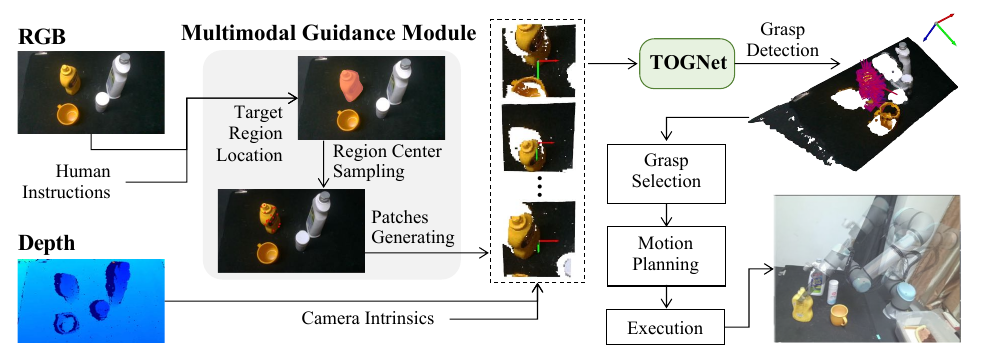}
\caption{\textbf{System Overview}. Taking a monocular RGB-D image as input, the \textit{\textbf{Multimodal Guidance Module (MGM)}} processes various types of guidance (e.g., language instructions, pointing gestures, and clicks) to locate target regions and sample points as region centers. The neighboring points around the centers are then clustered into multiple local patches. \textit{TOGNet} then extracts geometric features, predicts grasps locally and transforms them back to the original scene.}
\label{fig:framework} 
\vspace{-0.5cm} % adjust distance above distance 
\end{figure}

As shown in Fig. \ref{fig:framework}, our target-oriented grasping system comprises of two main components: \textit{\textbf{Multimodal Guidance Module (MGM)}} and \textit{\textbf{T}}arget-\textit{\textbf{O}}riented \textit{\textbf{G}}rasp \textit{\textbf{Net}}work (\textit{\textbf{TOGNet}}). \textit{MGM} leverages multimodal guidance methods to sample potential region centers. These centers are subsequently clustered with neighboring points to form multiple distinct patches. \textit{TOGNet} then extracts local geometric features from these patches and predicts grasps aligned with corresponding region centers by employing three specially designed heads. The framework's region-focal methodology facilitates the integration of various guidance methods and significantly enhances grasp quality, especially in unseen scenarios. \textit{TOGNet}'s proficiency in identifying grasp-related and object-agnostic geometric features from regional points is a critical factor in this enhanced performance.

\subsection{Multimodal Guidance Module}

The \textit{\textbf{Multimodal Guidance Module (MGM)}} is engineered to receive various types of human guidance and transform them into target-referenced local patches. The primary challenges include the precise recognition of instructions and accurate cropping of the target object. In this paper, we explore three types of guidance modes: language instructions, pointing gestures, and interactive clicks. These three modes are common in numerous interaction scenarios for target object grasping and are particularly beneficial for individuals with impairments. For instance, blind individuals can interact with the system using language, while those with speech and hearing impairments can guide the system through pointing gestures or interactive clicks. 

\subsubsection{Language Instructions} 
The advent of foundation models such as the Segmentation Anything Model (SAM) \cite{kirillov2023segment} and its variants \cite{zhao2023fast, zhang2023faster} have showcased impressive zero-shot performance in generating segmentation masks from box or point prompts. Leveraging Grounding DINO \cite{liu2023grounding}, which generates bounding boxes from text input, we further harness Grounded SAM \cite{ren2024grounded} with the replaced MobileSAM \cite{zhang2023faster} for its efficiency. This setup allows processing language inputs like ``Give me the yellow bottle" to produce segmentation masks from RGB images, as illustrated in Fig. \ref{fig:teaser}. Subsequently, several region centers are randomly selected from the mask to crop patches containing the target region.

\subsubsection{Pointing Gestures} 
Pointing gestures have been widely used to improve human-robot interaction \cite{constantin2022interactive, constantin2023multimodal}. MediaPipe Hand Landmarker \cite{lugaresi2019mediapipe} is an efficient method to detect hand landmarks (key points of the hands) from RGB images. We utilize the tool by first estimating the key points of the hand and then extracting the four key points of the index finger. Then, a 3D line is fitted from the key points using SVD \cite{sorkine2017least}. The intersection of this line with the 3D scene point cloud identifies the target object, from which we sample several neighboring points as region center points for cropping.

\subsubsection{Interactive Clicks} 
Interaction through clicks involves capturing the pixel coordinates on an ego-centric RGB image using the OpenCV library \cite{howse2013opencv}. These coordinates are transformed into 3D space and indicate the target's position. Ultimately, we sample region center points from the neighboring points.

Overall, individuals can specify the target object using language, pointing gestures, or clicks on the images. Our \textit{MGM} is designed to be extensible, potentially incorporating additional guidance like gaze tracking or voice commands by integrating more related sensors and advanced recognition algorithms.

\subsection{Target-Oriented Grasp Network} \label{section_TOGNet}

\begin{figure}[t]
\centering
{\includegraphics[width=0.90\linewidth]{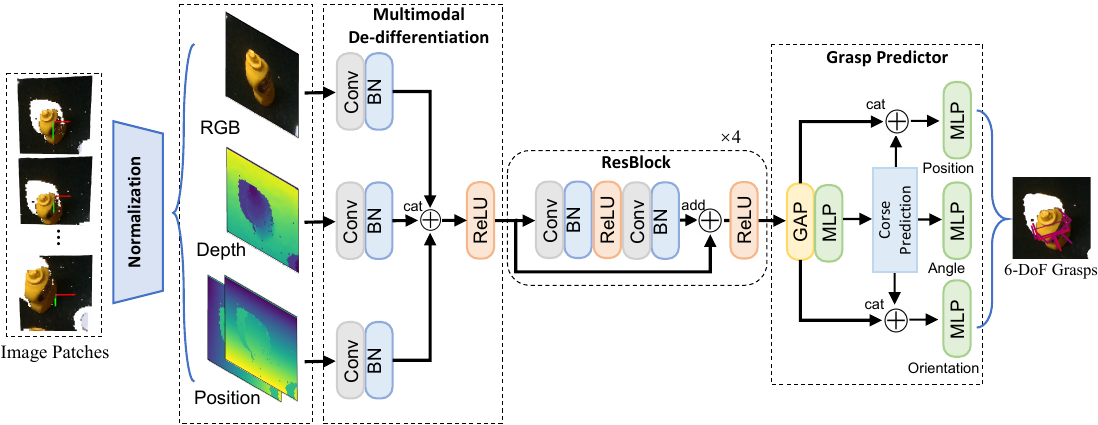}}
\captionof{figure}{Detailed structure of proposed \textit{TOGNet}.}
\label{fig:TOGNet} 
\vspace{-0.5cm} % adjust distance above distance 
\end{figure}

Traditional scene-level methods \cite{zhao2021regnet,fang2020graspnet,wang2021graspness,liu2022transgrasp,chen2023hggd} process the scene-level inputs and then adopt filters to obtain target-oriented grasps. Although such a scene-level grasp detection pipeline is effective for general grasp detection, when faced with the task of target-oriented grasping, it introduces unnecessary computation costs for non-targeted parts and may struggle to generate feasible grasps on the target object in clutters. Conversely, our method exclusively focuses on patch processing and region-focal grasp detection, in which only those grasps near the patch center are considered. In this regard, we develop the \textit{\textbf{T}}arget-\textit{\textbf{O}}riented \textit{\textbf{G}}rasp \textit{\textbf{Net}}work (\textit{\textbf{TOGNet}}), a robust network designed for extracting local features in the target patches and predicting grasps. Our \textit{TOGNet} is capable of predicting high-quality grasps solely utilizing local information. This advanced capability aligns seamlessly with the functionalities of our \textit{MGM}. As illustrated in Fig. \ref{fig:TOGNet}, our proposed \textit{TOGNet} consists of three parts: Multimodal De-differentiation, ResBlock, and Grasp Predictor.

\begin{figure}[t]
\centering
{\includegraphics[width=0.85\linewidth]{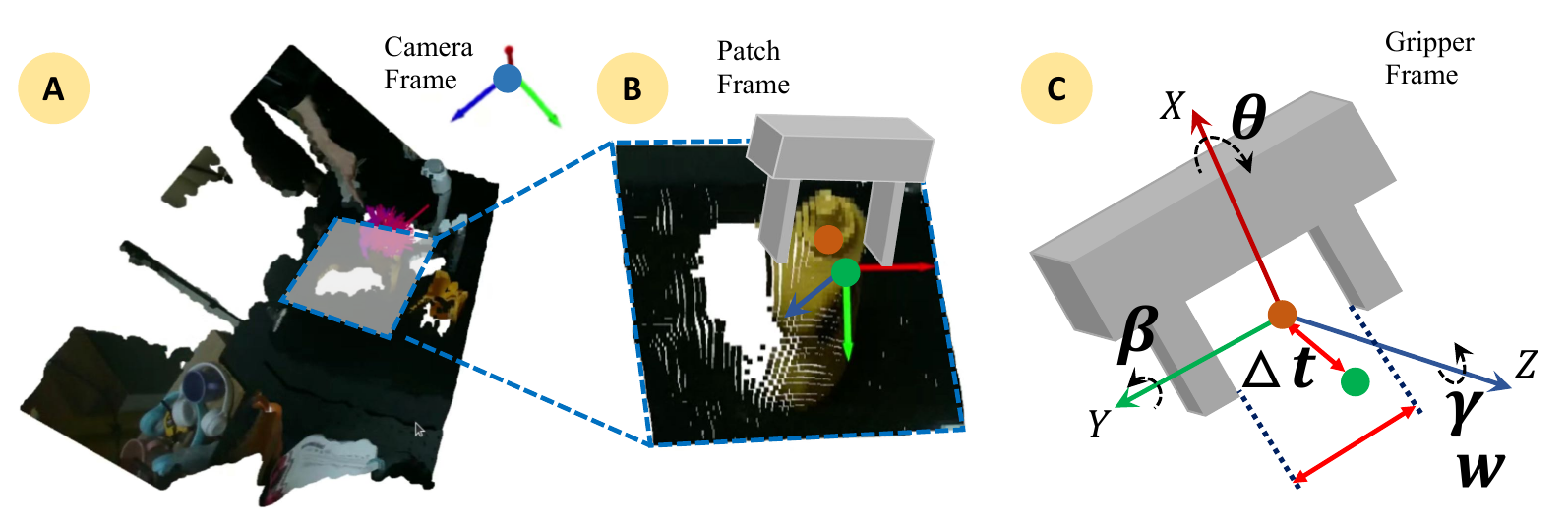}}
\caption{\textbf{Problem formulation of region-focal grasp detection.} A: The patch is cropped from the input RGB-D image. B: The local patches are transformed and normalized. C: The regional grasp representation as $(\Delta \mathbf{t}, \theta, \beta, \gamma, w)$.}
\label{fig:representation} 
% \vspace{-0.5cm} % adjust distance above distance 
\end{figure}

\subsubsection{Region-focal Grasp Detection} \label{section_regionfocal}

Different from the previous scene-level grasp detection approaches \cite{morrison2018closing,fang2020graspnet,chen2023hggd}, we reformulate the problem from a \textbf{region-focal viewpoint}, which focus on generating grasps close enough to the region center and under the grasp coverage threshold proposed in \cite{mousavian20196, sundermeyer2021contact}. Assuming multiple RGB-D region patches have been extracted from the guidance module, we aim to predict 6-DoF grasps:
\begin{equation}
    \boldsymbol{G_p} = \Phi(f_{i} |i=1,...,K),
\end{equation}
where $f_i \in R^{N \times 3}$ is the $i$-th region patch which is cropped from the RGB-D image and centered at $(x_p,y_p,z_p)$ in the camera frame, as shown in Fig. \ref{fig:representation}(A). And $\Phi(\cdot)$ denotes \textit{TOGNet} which predicts grasps $\boldsymbol{g_p} \in \boldsymbol{G_p}$ centered at $(x_p,y_p,z_p)$:
\begin{equation}
    \boldsymbol{g_p} = (\Delta \mathbf{t}, \theta, \beta, \gamma,  w).
\end{equation}
As shown in Fig. \ref{fig:representation}(C), $(\theta,\beta,\gamma)\in [-\frac{\pi}{2}, \frac{\pi}{2}]$ are grasp Euler angles in the gripper frame. $\theta$ represents the gripper in-plane rotation and $\beta,\gamma$ represents the gripper orientation. To avoid collision in clutters, $w$ denoting the grasp width is also predicted. Given that the guidance may be imprecise, to improve the robustness and accuracy, we introduce a 3D position offset, $\Delta \boldsymbol{t}=(\Delta x, \Delta y, \Delta z)\in [-2\ \text{cm}, 2\ \text{cm}]$, to finetune grasp position. Once $\boldsymbol{g_p}$ has been determined, the final grasp $\boldsymbol{g}$ on the target object can be represented as:
\begin{equation}
    \boldsymbol{g} = (x_p+\Delta x, y_p+\Delta y, z_p+\Delta z, \theta, \beta, \gamma, w).
\end{equation}

\subsubsection{Multimodal De-differentiation} 

The quality of the depth image is greatly influenced by sensor noise, especially on the edges, while RGB images are more resistant to noise and provide clearer object edges. Therefore, for robust and high-quality grasp detection, it is important to incorporate RGB information alongside the depth images. Moreover, since grasp poses are also related to object position, we include patch-level positional information, namely the X and Y coordinates of each pixel in the RGB-D images calculated from the camera intrinsics, enabling the network to capture the positional relationship between grasps and objects.

However, we have noticed that features extracted from different modalities exhibit significantly different distributions. Taking inspiration from channel de-differentiation \cite{yang2020cn} , we have investigated methods for aligning the modalities to process diverse inputs uniformly and ensure stable model training. In our approach, we implement multimodal alignment by applying a single-layer convolution and batch normalization before the feature extraction process:
\begin{equation}
    f=\text{ReLU}(\text{concat}(\text{BN}(\text{Conv}(f_{rgb})),\text{BN}(\text{Conv}(f_{depth})),\text{BN}(\text{Conv}(f_{position})))),
\end{equation}
where Conv is one-layer convolution, and BN means batch normalization. The ReLU activation function is applied after features of modalities are concatenated along the channel axis. By conducting modality normalization before the neural networks, we are able to process inputs with different modalities equally and train our network more stably. This minor improvement allows us to seamlessly incorporate RGB and positional data into our model, thus mitigating dimension mismatch and ensuring robust model training.

\subsubsection{Grasp Predictor}  

After aligning the multimodal inputs, we adopt a lightweight ResNet-18 to extract patch-level features efficiently. With the local features extracted, \textit{TOGNet} employs a two-stage coarse-to-fine paradigm to predict region-focal grasps. Initially, the extracted features after Global Average Pooling (GAP) are processed by a Multilayer Perceptron (MLP) to generate grasp-related features and coarse grasp attributes. Then, in the second stage, the geometric features are concatenated and then fed into three specialized MLP heads – \textbf{\textit{Position Head}}, \textbf{\textit{Angle Head}}, and \textbf{\textit{Orientation Head}} to predict the final fine grasp attributes.

In the \textbf{\textit{Position Head}}, we approach position and width prediction as regression problems to prevent collisions with nearby objects and refine grasp placements within each region. In the \textbf{\textit{Angle Head}}, we predict the grasp in-plane rotation angle $\theta \in [-\frac{\pi}{2}, \frac{\pi}{2}]$ by discretizing $\theta$ into $6$ bins and conducting bin classification and residual regression. In the \textbf{\textit{Orientation Head}},  following the non-uniform anchor sampling strategy \cite{chen2023hggd}, our strategy views spatial rotation prediction as a multi-label classification task. For fair comparisons, we keep the detailed settings the same as that in \cite{chen2023hggd}. With $7$ anchors each for $(\beta, \gamma)$, we generate up to $N_A=7^2=49$ possible grasps per region and preserve those with the highest scores. 

\subsection{Region-focal Dataset Generation} \label{section_data}

To effectively train \textit{TOGNet}, we generate a new dataset of regional RGB-D patches with grasp labels derived from the GraspNet-1Billion \cite{fang2020graspnet} dataset. The dataset creation involves selecting potential grasp centers and cropping local regions around them. Intuitively, randomly sampling grasp center candidates across the entire space seems straightforward but is inefficient, resulting in a high proportion of invalid data. Due to domain shifts, sampling centers directly from the projected grasp labels also lead to suboptimal results during inference.

\begin{figure}[t]
% \vspace{-0.8cm} % adjust distance above distance 
\centering
{\includegraphics[width=0.97\textwidth]{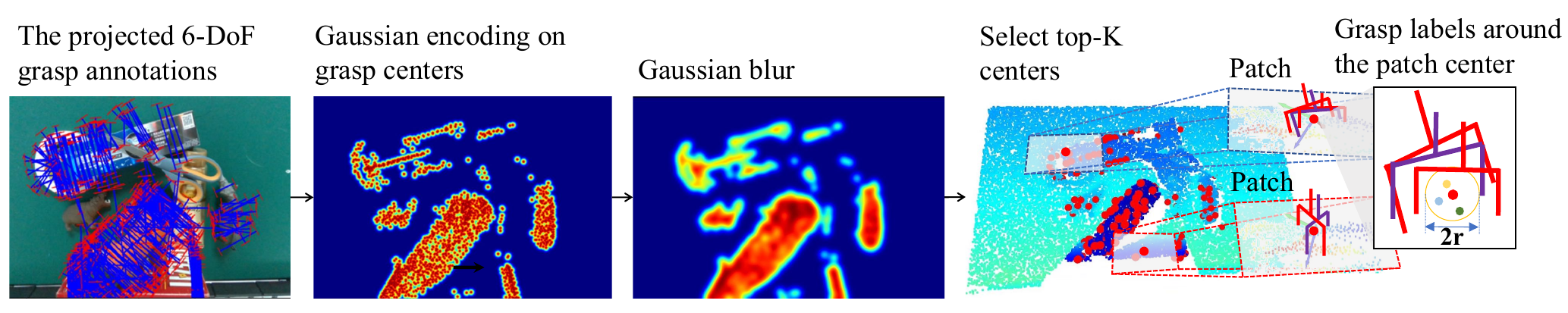}}
\caption{The pipeline of region-focal dataset generation. Grasp centers are sampled using the Gaussian-based strategy. Then, local neighboring points around each center are cropped as patches. Only the grasp labels within a radius of $r$ from the patch center are preserved.}
\label{fig:datageneration}
\vspace{-0.5cm} % adjust distance above distance 
\end{figure}

To address these challenges, we adopt a Gaussian-based strategy for center selection, which introduces a small proportion of invalid data and naturally incorporates noise into the training process. As illustrated in Fig. \ref{fig:datageneration}, the pipeline begins by projecting 6-DoF grasp annotations onto a 2D plane as 4-DoF planar grasps. We then encode each planar grasp center with a Gaussian kernel. Then, another Gaussian filter is applied to generate a blurred heatmap representing the distribution of graspable areas. Centers are selected from this heatmap via grid-based sampling and transformed into 3D points using the corresponding depth values.

Consistent with the grasp coverage criterion in \cite{mousavian20196, sundermeyer2021contact, chen2023hggd}, label processing involves preserving only those grasp labels within a $r=2$ cm Euclidean distance from the selected region centers. This approach emphasizes \textit{TOGNet}'s region-focal nature and ensures it focuses on generating grasps near regional centers, which is crucial for target-oriented grasping. To facilitate network training, points and labels are transformed from the camera frame to the local region frame (we translate the origin to each patch center), normalizing the data distribution.

Through the above steps, we have constructed a substantial dataset comprising millions of region-focal and object-agnostic regional data, providing a robust foundation for training \textit{TOGNet}.

\subsection{Losses} \label{implementation}

As depicted in Section. \ref{section_TOGNet}, the model incorporates loss components of the prediction heads. Thus, the overall loss can be formulated with different loss terms as
\begin{equation}
    L = L_{angle\_cls} + L_{angle\_reg} + L_{orientation} + L_{offset} + L_{width},
\end{equation}
where $L_{angle\_cls}$ and $L_{angle\_reg}$ represent the classification and residual regression loss for gripper in-plane rotation angle $\theta$ prediction. $L_{orientation}$ represents the focal loss \cite{lin2017focal} for the gripper orientation classification. $L_{width}$ and $L_{offset}$ are Smooth $L1$ losses employed to regress grasp width and center offset. 

\section{EXPERIMENT}

\subsection{Datasets and Metrics}

To fairly compare target-oriented grasp detection performance with other methods, we adopt the widely used GraspNet-1Billion dataset \cite{fang2020graspnet} as a standard evaluation platform, which contains RGB-D images captured in the real world from 190 cluttered scenes and more than 1 billion grasp annotations. We utilize the method in Section. \ref{section_data} to generate the region-focal data based on the GraspNet-1Billion dataset. Around \textbf{6.5M} local patches are obtained from the training split to train the proposed TOGNet.

The \textbf{Average Precision (AP)} \cite{fang2020graspnet} evaluation metric is commonly used in previous research to evaluate grasp quality, which leverages the force-closure scores \cite{qin2020s4g} of the top 50 grasps after non-maximum suppression to calculate the scene-level grasp detection quality. However, considering target-oriented grasping, only grasps on the target object are valid. Therefore, based on existing  evaluation metrics, we propose \textbf{Target-oriented Average Precision} to evaluate target-oriented grasp detection algorithms. First, we assign a specific target for each RGB-D image in the dataset by random sampling from objects not fully occluded by others. During evaluation, we measure the distance from detected grasp centers to the object mesh model and only retain the grasps on the correct objects. Consequently, due to the reduced grasp diversity, we compute force-closure scores \cite{qin2020s4g} for the \textbf{top 10} grasps after non-maximum suppression and utilize scores under different friction coefficients to derive Target-oriented Average Precision.

Furthermore, for the fair comparisons, the ground truth segmentation mask provided in the dataset is utilized for the purpose of scene-level grasps filtering (for scene-level methods) or guide patch generation (for \textit{TOGNet}).

\subsubsection{Quantitative Results}

\begin{table}[t]
\small
\renewcommand{\arraystretch}{1.25}
\caption{Target-oriented APs on GraspNet Dataset (RealSense/Kinect)}
\vspace{-0.5cm}
\begin{center}
\begin{tabular}{c|>{\centering\arraybackslash}p{2.15cm}>{\centering\arraybackslash}p{2.15cm}>{\centering\arraybackslash}p{2.15cm}|>{\centering\arraybackslash}p{2.15cm}} 
\toprule
\textbf{Method}&{\textbf{Seen}}&{\textbf{Similar}}&{\textbf{Novel}} &{\textbf{Mean}}\\
\midrule
GraspNet-baseline \cite{fang2020graspnet} & 22.64 / 13.28 & 20.63 / 12.67 & 8.35 / 3.85 & 17.21 / \;9.93\;\\
SBGrasp \cite{ma2023towards} & 38.04 / \quad-\quad\quad\ & 33.94 / \quad-\quad\quad\ & 15.97 / \quad-\quad\quad\ & 29.32 / \quad-\quad\quad\ \\
HGGD \cite{chen2023hggd} & 38.91 / 34.82 & 34.70 / 28.77 & 16.73 / 12.02 & 30.11 / 25.20\\
GSNet \cite{wang2021graspness} & 44.80 / 34.81 & 37.67 / 29.61 & 18.06 / 12.82 & 33.51 / 25.75\\
\midrule
TOGNet & \textbf{51.84} / \textbf{49.60} & \textbf{46.62} / \textbf{40.03} & \textbf{23.74} / \textbf{19.58} & \textbf{40.63} / \textbf{36.40}\\
\bottomrule
\end{tabular}
\label{graspnet}
\end{center}
\vspace{-0.2cm}

``-'': Result unavailable
\end{table}

We compare \textit{TOGNet}'s Target-oriented Average Precision with former SOTAs, which are also trained on the GraspNet-1Billion dataset.
Table \ref{graspnet} indicates that \textit{TOGNet} performs excellently on RGB-D images captured from RealSense and Kinect cameras. Compared with other methods, \textit{TOGNet} outperforms current state-of-the-art by a large margin on all dataset splits, proving the efficacy of our proposed multimodal region-focal neural network.

\subsubsection{Qualitative Results} 
\begin{figure}[t]
\centering
{\includegraphics[width=0.97\linewidth]{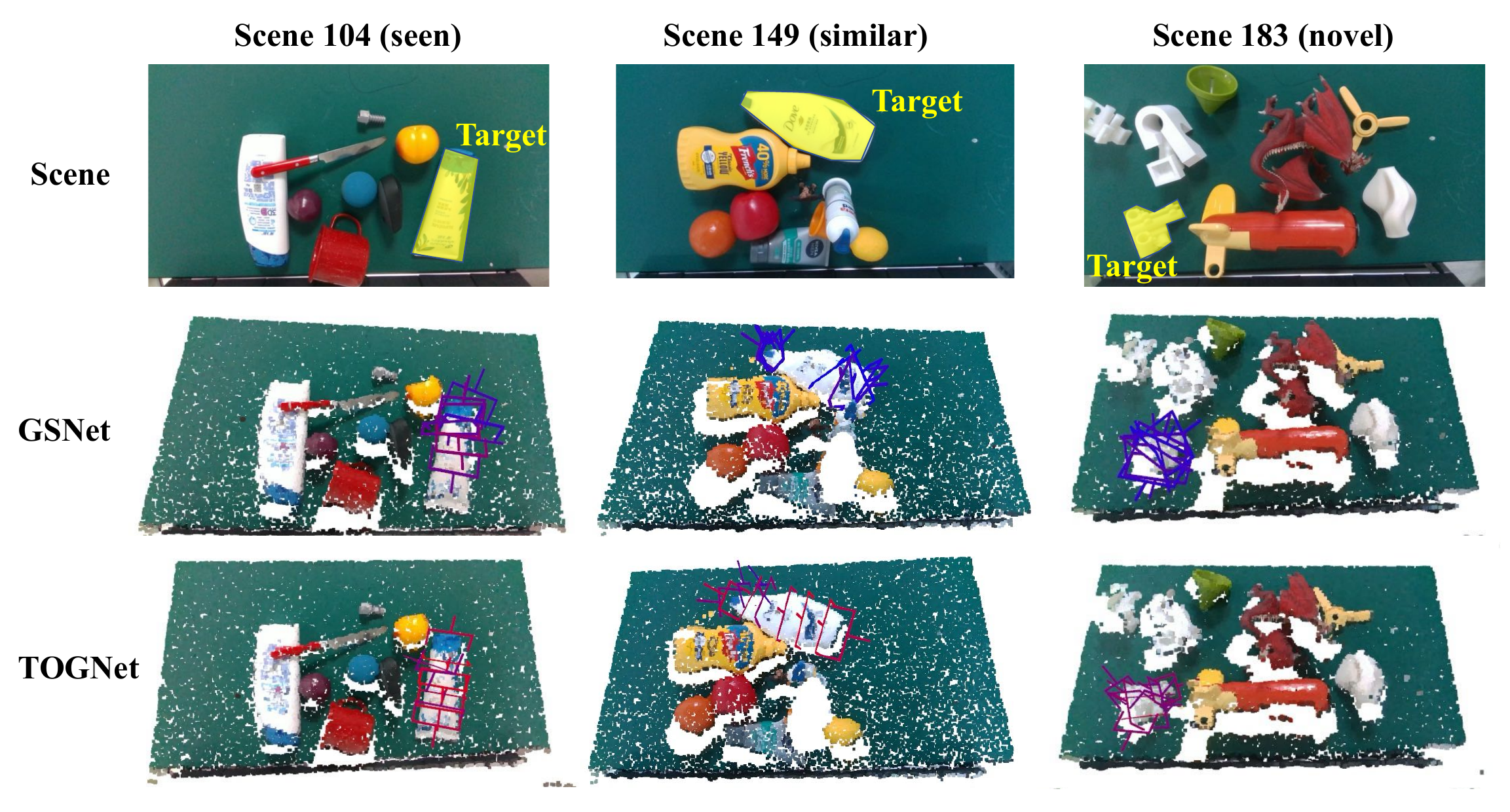}}
\captionof{figure}{\textbf{Qualitative results covering seen/similar/novel test set.} Top 10 grasps on the target objecs are displayed. Color implies the predicted grasp confidence (red: high, blue: low).}
\label{fig:quality} 
\end{figure}

We conduct grasp visualization of \textit{TOGNet} and GSNet \cite{wang2021graspness} on three parts (seen objects, similar objects and novel objects) of the test dataset. As depicted in Fig. \ref{fig:quality}, \textit{TOGNet} exhibits superior grasp detection quality and achieves a higher grasp coverage rate on the target objects than the former SOTA GSNet. This enhancement gives the robot more feasible action alternatives, benefiting scene-level grasp planning and execution. Furthermore, it is noteworthy that \textit{TOGNet} enables the generation of grasps that are better aligned with object centers and surfaces, avoiding possible collisions in cluttered scenes. The overall qualitative results underscore the validity of our algorithm for target-oriented grasping.

\begin{table}[t]
\centering
\small
\renewcommand{\arraystretch}{1.3}
\caption{Ablation Experiments, showing Target-oriented APs on RealSense Split.} 
\vspace{-3mm}
\setlength\tabcolsep{4px}
\begin{tabular}{ccc|ccc|c}
\toprule
\textbf{RF} & \textbf{PI} & \textbf{CI} & {\textbf{Seen}}&{\textbf{Similar}}&{\textbf{Novel}} &{\textbf{Mean}}\  \\ 
\midrule
  &   &   & 44.65 & 42.20 & 18.66 & 35.17 \\
\checkmark &   &   & 50.05 & 43.39 & 21.86 & 38.43  \\
\checkmark & \checkmark &   & 51.48 & 44.68 & 23.37 & 39.84 \\
\checkmark & \checkmark & \checkmark & \textbf{51.84} & \textbf{46.62}  & \textbf{23.74} & \textbf{40.63}  \\ 
\bottomrule
\end{tabular}
\label{ablation_model1}
\\
\vspace{0.3cm}
\textbf{RF}: Region-Focal,  \textbf{PI}: Positional Information,  \textbf{CI}: Color Information.  
\vspace{-0.5cm} % adjust distance above distance 
\end{table}

\subsubsection{Ablation Studies}

Table \ref{ablation_model1} presents the ablation results conducted on the \textit{TOGNet}. We first conduct ablation of our region-focal grasp detection setting by increasing the distance threshold in Section \ref{section_regionfocal} from $2$ cm to $4$ cm, which allows \textit{TOGNet} to generate grasps farther from the patch center. As the results show, keeping the region-focal grasp detection paradigm improves performance significantly. Constructing multimodal input by introducing XY coordinate maps (Positional Information) and RGB maps (Color Information) leads to considerable improvements. While geometric information from Z maps plays a crucial role in 6-DoF grasp detection, the additional positional information from XY maps and the inclusion of colors contribute to enhanced feature extraction. They are especially beneficial for generalizing unseen (similar and novel) scenes.

\subsection{Simulation Experiments}

\begin{figure}[t]
% \vspace{-0.8cm} % adjust distance above distance 
\centering
\includegraphics[width=12cm]{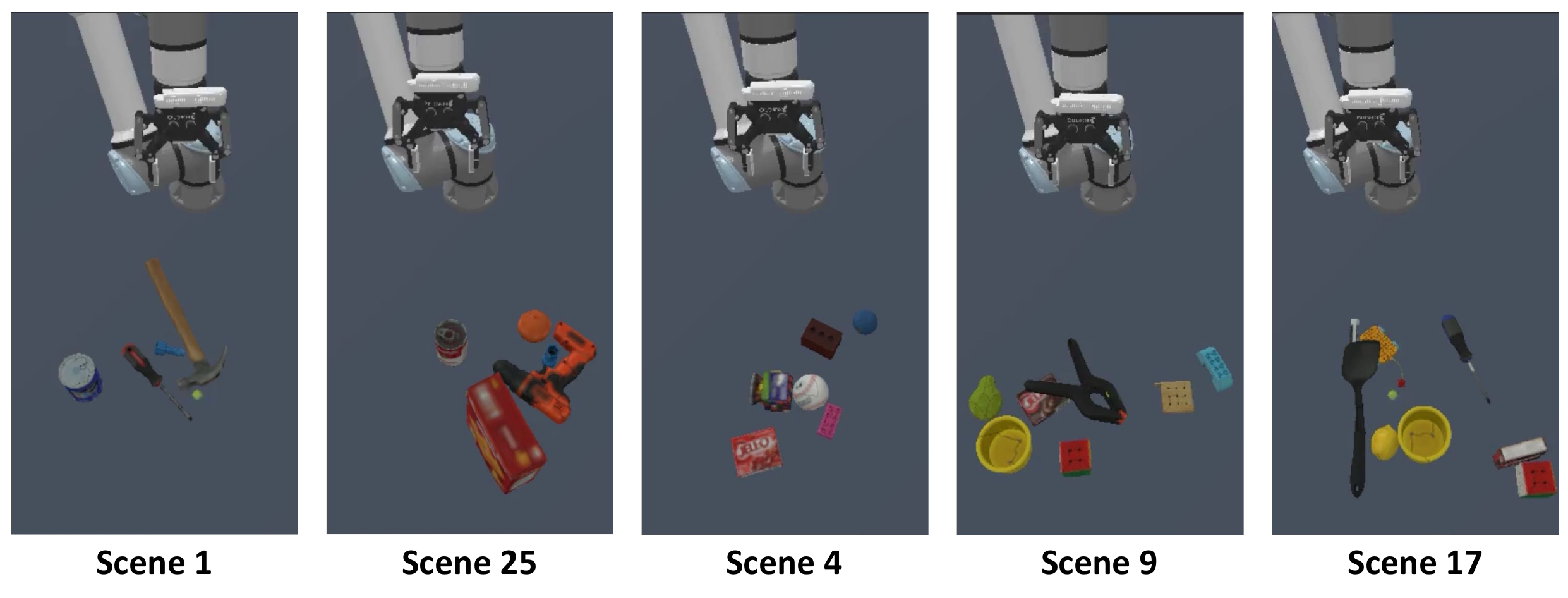}
\caption{Simulation experimental Settings. We present exemplar scenes from a total of 50 simulated scenes. In these scenes, the number of objects ranges from 4 to 8.}
\label{fig:sim}
% \vspace{-0.5cm} % adjust distance above distance 
\end{figure}

To assess the actual performance of \textit{TOGNet} for target-oriented grasping, we leverage the \textit{PickClutterYCB} environment within the Maniskill2 \cite{gu2023maniskill2} benchmark, modifying certain settings to suit our specific requirements. We set up a UR5e robotic arm with a Robotiq 2F-85 gripper and a wrist-mounted Realsense RGB-D camera, mirroring our real-world platform configuration. As depicted in Fig. \ref{fig:sim}, we select the first 50 clutter scenes from Maniskill2, each containing between 4 and 8 objects. In each episode, we set each object as a target sequentially. We utilize the ground truth object mask as guidance to exclude the influence of guidance modules and concentrate solely on evaluating the performance of different grasp detectors.

\begin{table}[t]
\small
\renewcommand{\arraystretch}{1.3}
\setlength{\tabcolsep}{1mm}
\caption{Simulation experimental results on 50 scenes with 329 target-oriented grasping episodes.}
\vspace{-0.5cm}
\begin{center}
\begin{tabular}{>{\centering\arraybackslash}p{2.5cm}|>{\centering\arraybackslash}p{3.5cm}} 
\toprule
\textbf{Method} & \textbf{Success Rate}\\
\midrule
GSNet \cite{wang2021graspness} & 165 / 329 = 50.2\%\\
HGGD \cite{chen2023hggd} & 178 / 329 = 54.1\%\\
AnyGrasp \cite{fang2023anygrasp} & 198 / 329 = 60.2\%\\
\midrule
TOGNet & \textbf{243} / 329 = \textbf{73.9}\%  \\
% \textbf{Success Rate}
% & \makecell[c]{165/329 \\ 50.2\%} & \makecell[c]{178/329 \\ 54.1\%} & \makecell[c]{198/329 \\ 60.2\%} & \makecell[c]{\textbf{243/329} \\\textbf{73.9}\%} \\
\bottomrule
\end{tabular}
\label{tab:sim}
\vspace{-0.5cm} % adjust distance above distance 
\end{center}
\end{table}

In our experiments, we randomly sample several pixels from the target object mask and generate corresponding patches, which are then fed into \textit{TOGNet} to detect grasps around the target area. We sample multiple patches to encapsulate different parts of the object to ensure comprehensive coverage, generating many high-quality grasps. We choose GSNet \cite{wang2021graspness}, HGGD \cite{chen2023hggd}, and AnyGrasp \cite{fang2023anygrasp} as comparative baselines, noted for their exemplary performance. These approaches typically detect grasps for the entire scene; however, for our target-oriented setup, we adapt them by filtering out redundant grasps distant from the target object. Ultimately, we get multiple target-referenced grasps and choose the grasp with the highest score. The robot then plans and executes the grasp to remove the object from the clutter.

% \subsubsection{\textbf{Results}}
We compute the success rate of all episodes as the evaluation metric. As indicated in Table \ref{tab:sim}, our \textit{TOGNet} achieves 73.9\%, which improves about 13.7\% compared with the second best baseline AnyGrasp \cite{fang2023anygrasp}. For three baseline methods, we find most failures arise because the filtered grasps around the target object are not good (i.e., the high-quality grasps are clustered around the same object). In contrast, our model predicts grasps for the target object locally, significantly enhancing performance and efficiency.

\subsection{Real Robot Experiments}

\begin{figure}[t]
% \vspace{-0.8cm} % adjust distance above distance 
\centering
\includegraphics[width=12cm]{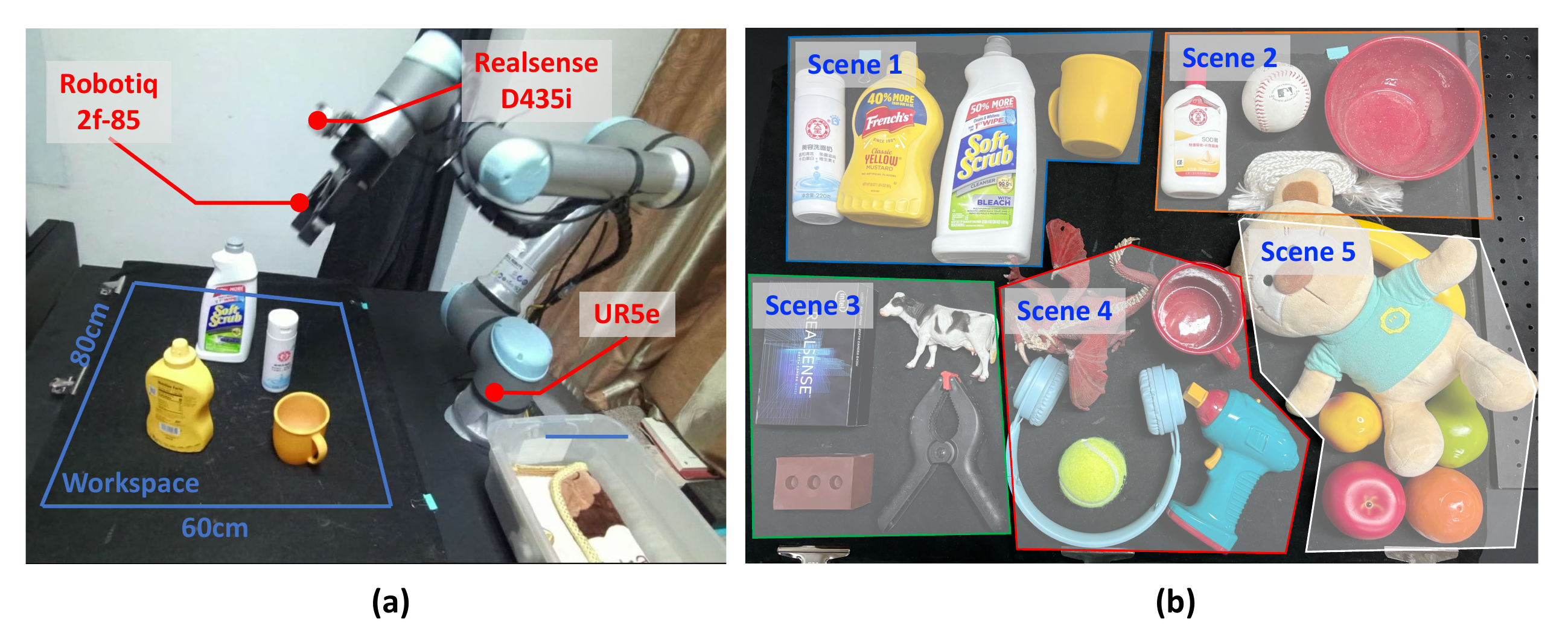}
\caption{Real-world robot experimental settings. (a) Several everyday objects are randomly placed in the workspace, and the robot captures visual information from the ego-centric RGB-D camera. (b) Real evaluated scenes: We set each object as a target for different guidance methods to assess the success rate.}
\label{fig:real}
\end{figure}
\begin{table}[t]
\renewcommand{\arraystretch}{1.3}
\setlength{\tabcolsep}{1mm}
\caption{Results of real robot experiments. We consider a trail successful if the target is removed from the workspace. }
\vspace{-0.5cm}
\begin{center}
\begin{tabular}{c|c c c c c c} 
\toprule
\textbf{Guidance} & \textbf{Scene1} & \textbf{Scene2} & \textbf{Scene3} & \textbf{Scene4} & \textbf{Scene5} & \textbf{Total}\\
\midrule
Language Instructions & 3/4 & 4/4 & 3/4 & 4/5 & 5/6 & 19/23 $\approx$ 82.6\% \\
Pointing Gesture & 4/4 & 4/4 & 3/4 & 4/5 & 5/6 & 21/23 $\approx$ 91.3\% \\
Interactive Click & 4/4 & 4/4 & 3/4 & 4/5 & 5/6 & 20/23 $\approx$ 87.0\% \\
\bottomrule
\end{tabular}
\\
% 1^1 \textbf{Success} / \textbf{Attempt} = \textbf{Success Rate} \\
% 2^2 \textbf{Cleared Scene} / \textbf{Total Scene} = \textbf{Completion Rate}
\label{tab:real}
\vspace{-0.5cm} % adjust distance above distance 
\end{center}
\end{table}

As illustrated in Fig. \ref{fig:real}, we perform real-world robotic grasping experiments using a UR-5e robot equipped with a Robotiq 2F-85 parallel-jaw gripper in a workspace measuring $80\times60$ cm. To capture ego-centric RGB-D images, we employ a Realsense-D435i camera.
Our experiment assembles a collection of 23 objects with diverse shapes and sizes commonly encountered in daily life, and we randomly select and arrange 4 to 6 of these objects to form 5 scenes, placing them in the workspace with various orientations. 

To evaluate the performance of our system, we demonstrate the language instructions, pointing gestures, and interactive click settings to conduct guidance. Please see the supplementary video for demonstration details. We adopt \textbf{Success Rate} as the evaluation metric. Table \ref{tab:real} reports that using language instructions and interactive clicks have an average grasp success rate of 82.6\% and 87.0\% across all scenes, while using the guidance of point gestures achieves the best grasp performance of a total success rate of 91.3\%. It indicates that our system generates high-quality grasps efficiently and performs well in the real world. Some failures are observed when the gripper slides over the object due to smooth surfaces, or sometimes the gripper collides with objects in the clutter.

\section{CONCLUSION}

Target-oriented grasping is pivotal in human-robot interaction, particularly in assisting individuals with disabilities in their daily activities. In this work, we propose a system that integrates a \textbf{\textit{Multimodal Guidance Module (MGM)}} and a \textbf{\textit{T}}arget-\textbf{\textit{O}}riented \textbf{\textit{G}}rasp \textbf{\textit{Net}}work (\textbf{\textit{TOGNet}}). Our system supports language instructions, pointing gestures, and interactive clicks to locate target objects accurately. \textit{TOGNet} processes these localized patches to efficiently generate high-quality grasps. Our model demonstrates superior performance from the modified GraspNet-1Billion Dataset specifically for target-oriented tasks. Additionally, \textit{TOGNet}'s efficacy is validated in a simulation benchmark that outperforms other state-of-the-art methods. Real-world robotic grasping experiments further confirm the effectiveness of our integrated system.

Despite these advancements, challenges remain. (1) segmentation results from Grounded SAM are not consistently accurate, occasionally producing multiple confusing or even wrong boxes that complicate subsequent processes. (2) Our system can not handle complex clutter scenarios like stacked, transparent, or reflective objects. (3) It cannot interpret complex instructions, such as ``Open the door" or ``Put the bottle in the container." By leveraging advanced VLMs, we can parse complex instructions and divide tasks into manageable stages to address complex tasks. Furthermore, it is crucial to implement assistance systems on some resource-constrained devices. Our TOGNet is efficient and easy to deploy on these devices because of its simple CNN structure, which enhances the overall user experience for future application.

% \clearpage\mbox{}Page \thepage\ of the manuscript.
% \clearpage\mbox{}Page \thepage\ of the manuscript.
% \clearpage\mbox{}Page \thepage\ of the manuscript.
% \clearpage\mbox{}Page \thepage\ of the manuscript.
% \clearpage\mbox{}Page \thepage\ of the manuscript. This is the last page.
% Now we have reached the maximum length of an ECCV \ECCVyear{} submission (excluding references).
% References should start immediately after the main text, but can continue past p.\ 14 if needed.
% \clearpage  % TODO REVIEW/FINAL: This \clearpage needs to be removed from both review and camera-ready versions.

\bibliographystyle{splncs04}
\bibliography{main}
\end{document}